\documentclass{article}
\usepackage{spconf,amsmath,graphicx}

\title{Highway Long Short-Term Memory RNNs for Distant Speech Recognition}
\name{Yu Zhang$^{1}$, Guoguo Chen$^{2}$, Dong Yu${^3}$, Kaisheng Yao$^3$, Sanjeev Khudanpur$^{2}$, James Glass$^{1}$\sthanks{Part of the work reported here was carried out during the 2015 Jelinek Memorial Summer Workshop on Speech and Language Technologies at the University of Washington, Seattle, and was supported by Johns Hopkins University via NSF Grant No IIS 1005411, and gifts from Google, Microsoft Research, Amazon, Mitsubishi Electric, and MERL.}
}
\address{$^1$MIT CSAIL \quad $^2$JHU CLSP \quad $^3$Microsoft Research\\
{\footnotesize \tt \{yzhang87,glass\}@mit.edu, \{guoguo,khudanpur\}@jhu.edu, \{dongyu, Kaisheng.YAO\}@microsoft.com}
}

\begin{document}
\ninept
\maketitle
\begin{abstract}
In this paper, we extend the deep long short-term memory (DLSTM) recurrent neural networks by introducing gated direct connections between memory cells in adjacent layers. These direct links, called highway connections, enable unimpeded information flow across different layers and thus alleviate the gradient vanishing problem when building deeper LSTMs. We further introduce the latency-controlled bidirectional LSTMs (BLSTMs) which can exploit the whole history while keeping the latency under control. Efficient algorithms are proposed to train these novel networks using both frame and sequence discriminative criteria. Experiments on the AMI distant speech recognition (DSR) task indicate that we can train deeper LSTMs and achieve better improvement from sequence training with highway LSTMs (HLSTMs). Our novel model obtains $43.9/47.7\%$ WER on AMI (SDM) dev and eval sets, outperforming all previous works. It beats the strong DNN and DLSTM baselines with $15.7\%$ and $5.3\%$ relative improvement respectively.
\end{abstract}

\section{Introduction}

Recently the deep neural network (DNN)-based acoustic models (AMs) greatly improved automatic speech recognition (ASR) accuracy on many tasks \cite{CD-DNN-HMM-Trans-Dahl+2012, DNN-SWB-seide+2011, hybrid, DNN-NoiseRobust-Seltzer+2013}. Further improvements were reported by using more advanced models such as convolutional neural networks (CNNs) \cite{Swietojanski:SPL14} and long short-term memory (LSTM) recurrent neural networks (RNNs) \cite{BLSTMTIMIT, BLSTMAM, LSTM-Sak+2014}.

Although these new techniques help to decrease the word error rate (WER) on distant speech recognition (DSR) \cite{NNSDM}, DSR remains a challenging task due to the reverberation and overlapping acoustic signals, even with sophisticated front-end processing techniques \cite{KumataniMR12,HainBDGGHHKLW12,StolckeDSP} and multi-pass decoding schemes.

In this paper, we explore more advanced back-end techniques for DSR. It is reported \cite{LSTM-Sak+2014} that deep LSTM (DLSTM) RNNs help improve generalization and often outperform single-layer LSTM RNNs. However, DLSTM RNNs are harder to train and slower to converge. In this paper, we extend DLSTM RNNs by introducing a gated direct connection between memory cells of adjacent layers. These direct links, called highway connections, provide a  path for information to flow between layers more directly without decay. It alleviates the gradient vanishing problem and enables DLSTM RNNs training with virtually arbitrary depth. 
Here, we refer to an LSTM RNN with highway connections as HLSTM RNN.

To further improve the performance, we also introduce the latency-controlled bidirectional LSTM (LC-BLSTM) RNNs. In the LC-BLSTM RNNs, the past history is fully exploited similar to that in the unidirectional LSTM RNNs. However, unlike the standard BLSTM RNNs which can start model evaluation only after seeing the whole utterance, the LC-BLSTM RNNs only look ahead for a fixed number of frames which limits the latency. The LC-BLSTM can be much more efficiently trained than the standard BLSTM without performance loss. It also trains and decodes faster than context-sensitive-chunk BLSTMs \cite{CZ_ref:msrabir} which can only access limited past and future context.

Our study is conducted on the AMI single distant microphone (SDM) setup. We compare the standard (B)LSTM RNNs and highway (B)LSTM RNNs with $3$ and $8$ layers. We show that the highway (B)LSTM RNNs with dropout applied to the highway connection significantly outperform the standard (B)LSTM RNNs. The high way connection helps to train deeper networks better and the highway (B)LSTM RNNs seem to benefit more from sequence discriminative training. Overall, our proposed model decreased WER by $15.7\%$ over DNNs, $14.4\%$ over CNNs \cite{CNNSDM}, and $5.3\%$ over DLSTM RNNs relatively. To our best knowledge, the $43.9/47.7\%$ WER we achieved on the AMI (SDM) dev and eval sets is the best results reported on this task.\footnote{The tools and scripts used to produce the results reported in this paper are publicly available as part of the CNTK toolkit\cite{CNTK}, and anyone with access to the data should be able to reproduce our results.}.

The rest of the paper is organized as follows. In Section \ref{sec:related} we briefly discuss related work. In Section \ref{sec:HLSTM} we introduce standard DLSTM RNNs, BLSTM RNNs, and the highway (B)LSTM RNNs. In Section \ref{CZ_sec:improvements} we describe LC-BLSTM and the way to train such models efficiently with both frame and sequence discriminative criteria. We summarize the experimental setup in Section \ref{exp_setup} and report experimental results in Section \ref{sec:results}. Conclusions
are reiterated in Section \ref{sec:conclusion}.

\section{Related Work}
\label{sec:related}
After developing the highway LSTMs independently we noticed that similar work has been done in \cite{highway, GatedLSTM, GridLSTM}. All of these works share the same idea of adding gated linear connections between different layers. The highway networks proposed in \cite{highway} adaptively carry some dimensions of the input directly to the output so that information can flow across layers much more easily. However, their formulation is different from ours and their focus is DNN. The work in \cite{GatedLSTM} share the same idea and model structure. \cite{GridLSTM} is more general and uses a generic form. However, their task is on text e.g. machine translation while our focus is distant speech recognition. In addition, we used dropout as the way to control the highway connections which turns out to be critical for DSR. 

\section{Highway Long Short-term Memory RNNs}
\label{sec:HLSTM}
\begin{figure*}[!ht]
\centering
\includegraphics[width=0.9\textwidth]{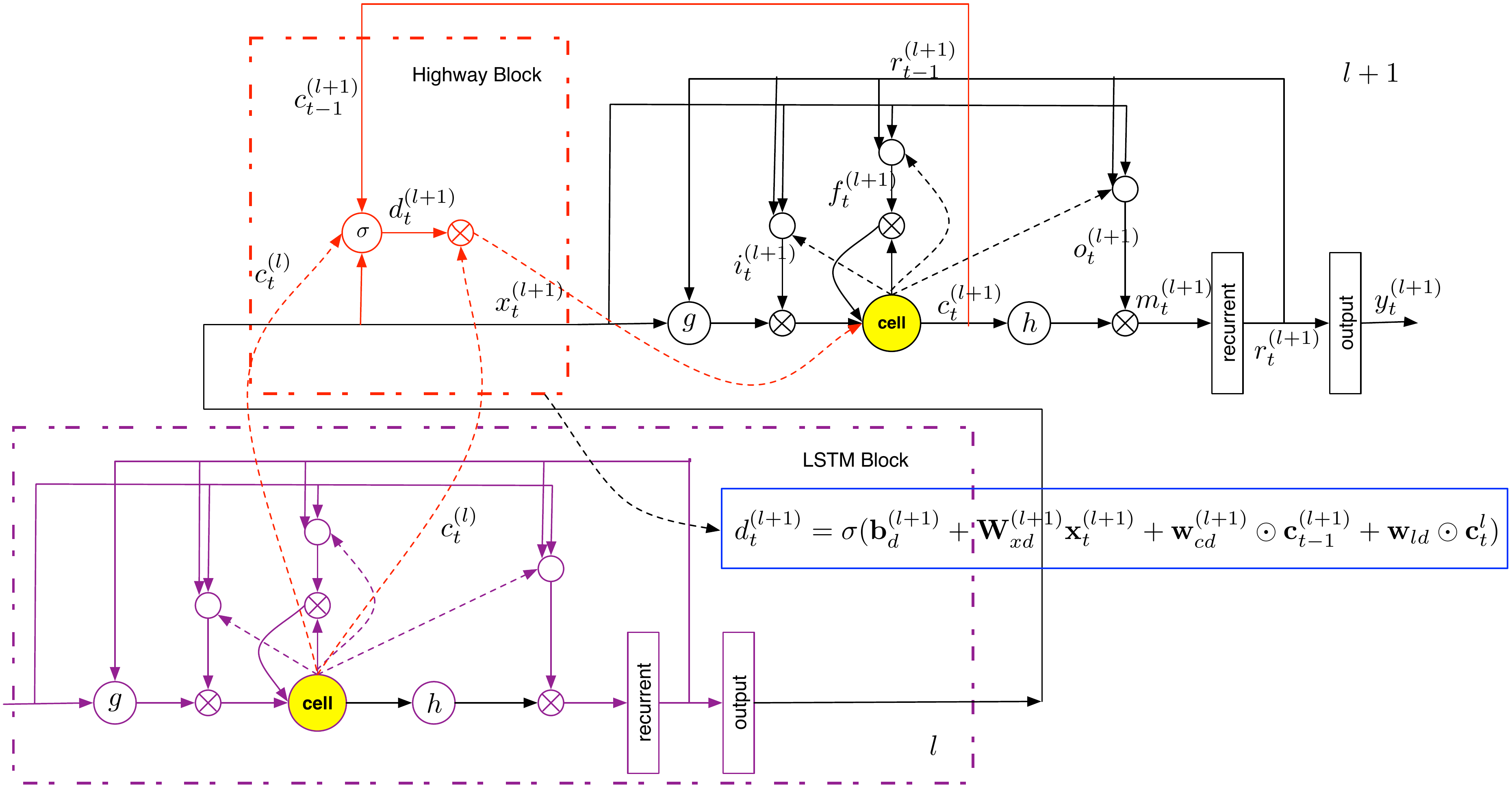}
  \vspace{-10pt}
\caption{Highway Long Short-Term Memory RNNs}
\label{fig:HLSTMP}
\end{figure*}
\subsection{Long short-term memory RNNs}
The LSTM RNN was initially proposed in \cite{LSTM} to solve the gradient diminishing problem in RNNs. It introduces a linear dependence between  $\mathbf{c}_t$, the memory cell state at time $t$, and $\mathbf{c}_{t-1}$, the same cell's state at $t-1$. Nonlinear gates are introduced to control the information flow. The operation of the network follows the equations
{\footnotesize
\begin{align}
    \mathbf{i}_t &= \sigma (\mathbf{W}_{xi}\mathbf{x}_t + \mathbf{W}_{mi}\mathbf{h}_{t-1} + \mathbf{W}_{ci}\mathbf{c}_{t-1}+\mathbf{b}_i)\label{eq:lstm1}\\
    \mathbf{f}_t &= \sigma (\mathbf{W}_{xf}\mathbf{x}_t + \mathbf{W}_{mf}\mathbf{h}_{t-1} + \mathbf{W}_{cf}\mathbf{c}_{t-1}+\mathbf{b}_f)\label{eq:lstm2}\\
    \mathbf{c}_t &= \mathbf{f}_t\odot \mathbf{c}_{t-1} + \mathbf{i}_t \odot \tanh (\mathbf{W}_{xc}\mathbf{x}_t + \mathbf{W}_{mc}\mathbf{m}_{t-1}+\mathbf{b}_c)\label{eq:lstm3}\\
    \mathbf{o}_t &= \sigma(\mathbf{W}_{xo}\mathbf{x}_t + \mathbf{W}_{mo}\mathbf{h}_{t-1}+\mathbf{W}_{co}\mathbf{c}_t + \mathbf{b}_o)\label{eq:lstm4}\\
    \mathbf{m}_t &= \mathbf{o}_t\odot \tanh(\mathbf{c}_t) \label{eq:lstm5}
\end{align}
}%
iteratively from $t=1$ to $T$, where $\sigma(\dot)$ is the logistic sigmoid function, $\mathbf{i}_t, \mathbf{f}_t,\mathbf{o}_t,\mathbf{c}_t$ and $\mathbf{m}_t$ are vectors to represent values at time $t$ of the input gate, forget gate, output gate, cell activation, and cell output activation respectively. $\odot$ denotes element-wise product of vectors. $\mathbf{W}_{*}$ are the weight matrices connecting different gates, and $\mathbf{b}_{*}$ are the corresponding bias vectors. All the matrices are full except the matrices $\mathbf{W}_{ci},\mathbf{W}_{cf},\mathbf{W}_{co}$ from the cell to gate vector which is diagonal.


\subsection{Deep LSTM RNNs}
Deep LSTM RNNs are formed by stacking multiple layers of LSTM cells. Specifically, the output of the lower layer LSTM cells $\mathbf{y}_t^l$ is fed to the upper layer as input $\mathbf{x}_t^{l+1}$. Although each LSTM layer is deep in time since it can be unrolled in time to become a feed-forward neural network in which each layer shares the same weights, deep LSTM RNNs still outperform single-layer LSTM RNNs significantly. It is conjectured \cite{LSTM-Sak+2014} that DLSTM RNNs can make better use of parameters by distributing them over the space through multiple layers. Note that in the conventional DLSTM RNNs the interaction between cells in different layers must go through the output-input connection. 

\subsection{HLSTM RNNs}
The Highway LSTM (HLSTM) RNN proposed in this paper is illustrated in Figure \ref{fig:HLSTMP}. It has a direct gated connection (in the red block) between the memory cells $\mathbf{c}_t^{l}$ in the lower layer $l$ and the memory cells $\mathbf{c}_t^{l+1}$ in the upper layer $l+1$. The carry gate controls how much information can flow from the lower-layer cells directly to the upper-layer cells. The gate function at layer $l+1$ at time $t$ is  
\begin{equation}
\footnotesize
    \mathbf{d}_t^{(l+1)} = \sigma (\mathbf{b}_d^{(l+1)} + \mathbf{W}_{xd}^{l+1}\mathbf{x}_t^{(l+1)} + \mathbf{w}_{cd}^{l+1}\odot \mathbf{c}_{t-1}^{(l+1)} + \mathbf{w}_{ld}^{(l+1)}\odot \mathbf{c}_t^{l}),
\end{equation}
where $\mathbf{b}_d^{(l+1)}$ is a bias term, $\mathbf{W}_{xd}^{(l+1)}$ is the weight matrix connecting the carry gate to the input of this layer. $\mathbf{w}_{cd}^{(L+1)}$ is a weight vector from the carry gate to the past cell state in the current layer. $\mathbf{w}_{ld}^{(L+1)}$ is a weight vector connecting the carry gate to the lower layer memory cell. $\mathbf{d}^{(l+1)}$ is the carry gate activation vectors at layer $l+1$. 

Using the carry gate, an HLSTM RNN computes the cell state at layer $(l+1)$ according to
{\footnotesize
\begin{align}
&\mathbf{c}_t^{l+1} = \mathbf{d}_t^{(l+1)}\odot \mathbf{c}_t^l + \mathbf{f}_t^{(l+1)}\odot \mathbf{c}_{t-1}^{(l+1)} \nonumber\\
&+ \mathbf{i}_t^{(l+1)}\odot \tanh(\mathbf{W}_{xc}^{(l+1)}\mathbf{x}_t^{(l+1)} + \mathbf{W}_{hc}^{(l+1)}\mathbf{m}_{t-1}^{(l+1)}+\mathbf{b}_c),
\end{align}
}%
while all other equations are the same as that in the standard LSTM RNNs as described in Eq. (\ref{eq:lstm1}),(\ref{eq:lstm2}),(\ref{eq:lstm4}), and (\ref{eq:lstm5}). 

Thus, depending on the output of the carry gates, the highway connection can smoothly vary its behavior between that of a plain LSTM layer or simply passes its cell memory from previous layer. The highway connection between cells in different layers makes influence from cells in one layer to the other more direct and can alleviate the gradient vanishing problem when training deeper LSTM RNNs.

\subsection{Bidirectional Highway LSTM RNNs}
The unidirectional LSTM RNNs we described above can only exploit past history. In speech recognition, however, future contexts also carry information and should be utilized to further enhance acoustic models. The bidirectional RNNs take advantage of both past and future contexts by processing the data in both directions with two separate hidden layers. It is shown in \cite{BLSTMTIMIT, BLSTMAM, CZ_ref:msrabir} that bidirectional LSTN RNNs can indeed improve the speech recognition results. In this study, we also extend the HLSTM RNNs from unidirection to bidirection. Note that the backward layer follows the same equations used in the forward layer except that $t-1$ is replaced by $t+1$ to exploit future frames and the model operates from $t=T$ to $1$. The output of the forward and backward layers are concatenated to form the input to the next layer.

\section{Efficient Network Training}
\label{CZ_sec:improvements}
Nowadays, GPUs are widely used in deep learning by leveraging massive parallel computations via
mini-batch based training. For unidirectional RNN models, to better utilize the parallelization
power of the GPU card, in \cite{CNTK}, multiple sequences (e.g., 40) are often packed into the
same mini-batch. Truncated BPTT is usually performed for parameters updating, therefore, only a
small segment (e.g., 20 frames) of each sequence has to be packed into the mini-batch. However, when applied to sequence level training (BLSTM or sequence training),  GPU's limited memory restricts
the number of sequences that can be packed into a mini-batch, especially for LVCSR tasks with long
training sequences and large model sizes. One alternative way to speed up is using asynchronous SGD
based on a GPU/CPU farm \cite{googleASGD}. In this section, we are more focused on fully utilizing the parallelization power of a single GPU Card. The algorithms proposed here can also be applied
to a multi-GPU setup.

\subsection{Latency-controlled bi-directional model training}
To speed up the training of bi-direcctional RNNs, the Context-sensitive-chunk BPTT (CSC-BPTT) is proposed in \cite{CZ_ref:msrabir}. In this method, a sequence is
firstly split into chunks of fixed length $N_c$. Then $N_l$ past frames and $N_r$ future frames are concatenated before and after each chunk as the left and right context, respectively. The appended frames
are only used to provide context information and do not generate error signals during training. Since each trunk can be independently drawn and trained, they can be stacked to form large minibatches to speed up training. 

Unfortunately, the model trained with CSC-BPTT is no longer the true bidirectional RNN since the history it can exploit is limited by the left and right context concatenated to the chunk. It also introduces additional computation cost during decoding since both the left and right contexts need to be recomputed for each chunk.

To solve the problems in the CSC-BPTT we propose the latency-controlled bi-directional RNNs. Different from the CSC-BPTT, in our new model we carry the whole past history while still using a truncated future context. Instead of concatenating and computing $N_l$ left contextual frames for each chunk we directly carry over the left contextual information from the previous chunk of the same utterance. For every chunk, both training and decoding computational cost is reduced by a factor of $\frac{N_l}{N_l+N_c+ N_r}$. Moreover, loading the history from previous mini-batch instead of a fixed contextual windows makes the context exact when compared to the uni-directional model. Note that the standard BLSTM RNNs come with significant latency since the model can only be evaluated after seeing the whole utterance. In the latency-controlled BLSTM RNNs the latency is limited to $N_r$ which can be set by the users. In our experiments, we process 40 utterances in parallel which is 10 times faster than processing the whole utterances without performance loss. Compared to the CSC BPTT our approach is 1.5 times faster and often leads to better accuracy.


\subsection{Two pass forward computation for sequence training}
\label{CZ_ssec:smbr_support}

To increase the number of sequences that can be fit into a mini-batch, we propose a
two-forward-pass solution for sequence discriminative training on top of recurrent neural
networks. The basic idea is pretty straightforward. For sequence 
training of recurrent models, we use the same mini-batch packaging method as that in 
the cross-entropy training case, i.e., we pack multiple sequences (e.g., 40) into the
same mini-batch, each with a small chunk (e.g., 20 frames). Then in the first 
forward pass, we collect log-likelihood of frames in the mini-batch, and put those 
into a pool, without updating the model. We do this until we have collected 
the log-likelihood for a certain number of sequences. At this point, we are able to 
compute the error signal for each of the sequence in the pool. We then
roll back the mini-batches that we have just computed error signals for, and start
the second forward pass, this time we update the model using the error signals from
the pool. With this two-forward-pass solution, we are able to pack far
more sequences in the same mini-batch, e.g., 40$\sim$60, thus
leading to much faster training.

\section{EXPERIMENT SETUP}
\label{exp_setup}
\subsection{Corpus}
We evaluated our models on the AMI meeting corpus \cite{AMI}. The AMI corpus 
comprises around 100 hours of meeting recordings, recorded in instrumented meeting
rooms. Multiple microphones were used, including individual headset microphones (IHM), lapel microphones, and one or more microphone arrays. In this work, we use the single distant microphone (SDM) condition for our experiments. Our systems are trained and tested using the split recommended in the corpus release: a training set of 80 hours, a 
development set and a test set each of 9 hours. For our training, we use all the segments provided by the corpus,
including those with overlapped speech. Our models are evaluated on the evaluation
set only. NIST's asclite tool \cite{asclite} is used for scoring.

\subsection{System description}
Kaldi \cite{kaldi} is used for feature
extraction, early stage triphone training as well as decoding. A maximum likelihood
acoustic training recipe is used to trains a GMM-HMM triphone system. Forced alignment is
performed on the training data by this triphone system to generate labels for further
neural network training.

The Computational Network Toolkit (CNTK) \cite{CNTK} is used for neural network
training. We start off by training a 6-layer DNN, with $2,048$ sigmoid units per layer.
40-dimensional filterbank features, together with their corresponding delta and 
delta-delta features are used as raw feature vectors. For our DNN training we concatenated
15 frames of raw feature vectors, which leads to a dimension of $1,800$. This DNN again is
used to force align the training data to generate labels for further LSTM training.

Our (H)LSTM models, unless explicitly stated otherwise, are added with a projection
layer on top of each layer's output, as proposed in \cite{LSTM-Sak+2014}, and are trained
with 80-dimensional log Mel filterbank (FBANK) features. For LSTMP models, each hidden layer
consists of 1024 memory cells together with a 512-node projection layer. For the BLSTMP
models, each hidden layer consists of 1024 memory cells (512 for forward and 512 for backward) with
a 300-node projection layer. Their highway companions share the same network structure,
except the additional highway connections.

All models are randomly initialized without either generative or discriminative pretraining
\cite{Feature-Eng-Seide+2011}. A validation set is used to control the learning rate which will
be halved when no gain is observed. To train the unidirectional model, the truncated
back-propagation-through-time (BPTT) \cite{BPTT} is used to update the model parameters. Each BPTT
segment contains 20 frames and process 40 utterances simultaneously. To train the latency-controlled bidirectional model, we set $N_c=22$ and $N_r=21$ and also process 40
utterances simultaneously. A start learning rate of 0.2 per minibatch is used and then the learning
rate scheduler takes action. For frame level cross-entropy training, $L2$ constraint regularization \cite{L2Reg} is used. For sequence training, $L2$ constraint regularization is also applied 
whenever it is used in the corresponding cross-entropy trained model. We use a fixed per sample
learning rate of $1e-5$ for DNN sequence training, and $2e-6$ for LSTM sequence training.

\section{RESULTS}
\label{sec:results}
The performance of various models are evaluated using word error rate
(WER) in percent below. All the experiments are conducted on AMI SDM1 
eval set, if not specified otherwise. Since we do not exclude the 
overlapping speech segments during model training, in addition to
results on the full eval set, we also show results on a subset that only
contains the non-overlapping speech segments as \cite{CNNSDM}.

\subsection{3-layer Highway (B)LSTMP}
\begin{table}[!ht]
\centering
\begin{tabular}{l|c|c|c}
\hline
 System   & \#Layers & with overlap & no overlap \\\hline\hline
 DNN      & 6 & 57.5     & 48.4 \\\hline\hline
 LSTMP    & 3 & 50.7     & 41.7 \\
 HLSTMP   & 3 & 50.4     & 41.2\\\hline\hline
 BLSTMP   & 3 & 48.5     & 38.9\\
 BHLSTMP  & 3 & 48.3     & 38.5\\\hline
\end{tabular}
\vspace{-7pt}
\caption{Performance of highway (B)LSTMP RNNs}
\label{tab:highway_3layer}
\end{table}
Table \ref{tab:highway_3layer} gives WER performance of the 3-layer
LSTMP and BLSTMP RNNs, as well as their highway versions. The
performance of the DNN network is also listed for comparison. From
the table, it's clear that the highway version of the LSTM RNNs
consistently outperform their non-highway companions, though with a
small margin.

\subsection{Highway (B)LSTMP with dropout}
\begin{table}[!ht]
\centering
\begin{tabular}{l|c|c|c}
\hline
 System           & \#Layers & with overlap & no overlap \\\hline\hline
 LSTMP            & 3 & 50.7     & 41.7 \\
 HLSTMP + dropout & 3 & 49.7     & 40.5 \\\hline\hline
 BLSTMP           & 3 & 48.5     & 38.9 \\
 BHLSTMP + dropout& 3 & 47.5     & 37.9  \\\hline
\end{tabular}
\vspace{-7pt}
\caption{Performance of highway (B)LSTMP RNNs with dropout}
\label{tab:highway_3layer_dropout}
\end{table}
Dropout is applied to the highway connection to control its flow: a high dropout rate
essentially turns off the highway connection, and a small dropout
rate, on the other hand, keeps the connection alive. In our
experiments, for early training stages, we use a small dropout rate of
$0.1$, and increase it to $0.8$ after 5 epochs of training. Performance
of highway (B)LSTMP networks with dropout is shown in
Table \ref{tab:highway_3layer_dropout}, as we can see, dropout helps
to further bring down the WER for highway networks.

\subsection{Deeper highway LSTMP}
\begin{table}[!ht]
\centering
\begin{tabular}{l|c|c|c}
\hline
 System & \#layers & with overlap & no overlap \\\hline\hline
 LSTMP  & 3 & 50.7 & 41.7 \\
 LSTMP  & 8 & 52.6 & 43.8 \\\hline\hline
 HLSTMP & 3 & 50.4 & 41.2 \\
 HLSTMP & 8 & 50.7 & 41.3 \\\hline
\end{tabular}
\vspace{-7pt}
\caption{Comparison of shallow and deep networks}
\label{tab:shallow_deep}
\end{table}
When a network goes deeper, the training usually becomes difficult.
Table \ref{tab:shallow_deep} compares the performance of shallow and
deep networks. From the table we can see that for a normal LSTMP
network, when it goes from 3 layers to 8 layers, the recognition
performance degrades dramatically. For the highway network, however,
the WER only increase a little bit. The table suggests that the
highway connection between LSTM layers allows the network to go
much deeper than the normal LSTM networks.

\subsection{Highway LSTMP with sequence training}
\begin{table}[!ht]
\centering
\begin{tabular}{l|c|c|c}
\hline
 System (dr: dropout)                        & \#Layers & with overlap      & no overlap \\\hline\hline
 DNN                      & 6 & 57.5          & 48.4 \\
 DNN  + sMBR              & 6 & 54.4          & 44.7 \\\hline\hline
 LSTMP                   & 3 & 50.7          & 41.7 \\
 LSTMP  + sMBR            & 3 & 49.3          & 39.8 \\
 HLSTMP                  & 3 & 50.4          & 41.2 \\
 HLSTMP  + sMBR           & 3 & 48.3          & 38.4 \\
 HLSTMP (dr) + sMBR & 3 & \textbf{47.7} & 38.2 \\\hline\hline
 LSTMP                     & 8 & 52.6          & 43.8 \\
 LSTMP  + sMBRR            & 8 & 50.5          & 41.3   \\
 HLSTMP                   & 8 & 50.7          & 41.3 \\
 HLSTMP + sMBR            & 8 & 47.9          & \textbf{37.7} \\\hline
\end{tabular}
\vspace{-7pt}
\caption{Performance of sequence training on various networks}
\label{tab:smbr}
\end{table}
We perform sequence discriminative training for the networks discussed in
Section \ref{CZ_ssec:smbr_support}. Detailed results are shown in Table \ref{tab:smbr}.
The table suggests that introducing the highway connection between LSTMP
layers is beneficial to sequence discriminative training. For example,
without the highway connection, sequence training on top of the 3-layer
LSTMP network brings WER from $50.7$ down to $49.3$, a relative 
improvement of only $3\%$ on this particular task. After introducing the 
highway connection and dropout, the improvement is from $50.4$ to $47.7$, $5\%$ relatively. The relative improvement is even larger on the
non-overlapping segment subset, which is roughly $7\%$. The results suggest that sequence training is beneficial from both highway connection and a deeper structure.

\section{CONCLUSION}
\label{sec:conclusion}
We presented a novel highway LSTM network, and applied it to a 
far-field speech recognition task. Experimental results suggest that this
type of network consistently outperforms the normal (B)LSTMP networks, especially when dropout is applied to the highway connection to control the connection's on/off state. Further experiments also
suggest that the highway connection allows the network to go much deeper,
and to get larger benefit from sequence discriminative training.

\vfill\pagebreak

\bibliographystyle{IEEEbib}
\bibliography{refs}

\end{document}